\documentclass[]{ceurart}

\usepackage[frozencache=true,cachedir=minted-cache]{minted}
\setminted{breaklines=true}

\usepackage{sidecap}
\usepackage{latexsym}
\usepackage{placeins}
\usepackage{subcaption}
\usepackage{enumitem}

\usepackage[T1]{fontenc}

\usepackage[utf8x]{inputenc}

\usepackage{microtype}

\usepackage{inconsolata}

\usepackage{graphicx}

\usepackage{hyperref}
\usepackage{url}
\usepackage{graphicx}
\usepackage{listings}
\usepackage{booktabs}       
\usepackage{multirow}

\DeclareUnicodeCharacter{2200}{$\forall$}
\DeclareUnicodeCharacter{65533}{$\rightarrow$}

\newtheorem{definition}{Definition}

\definecolor{backcolor}{rgb}{0.95,0.95,0.95}

\lstdefinestyle{mystyle}{
    backgroundcolor=\color{backcolor},
    breaklines=true,
    numbers=left,
    numbersep=5pt,
    basicstyle=\ttfamily\footnotesize,
    captionpos=b,
    escapeinside={\%*}{*)},
}
\begin{document}

\copyrightyear{2025}
\copyrightclause{Copyright for this paper by its authors. Use permitted under Creative
Commons License Attribution 4.0 International (CC BY 4.0).}
\conference{ Workshop on Knowledge Graphs and Neurosymbolic AI (KG-NeSy), co-located with SEMANTiCS’25: International Conference on Semantic Systems, September 3–5, 2025, Vienna, Austria}
%
%
\title{Intermediate Languages Matter: Formal Languages and LLMs affect Neurosymbolic Reasoning}


\author[1]{Alexander Beiser}[
    email=alexander.beiser@tuwien.ac.at]
\address[1]{TU Wien, Vienna, Austria}
\author[2]{David Penz}
\address[2]{Johannes Kepler University Linz, Linz, Austria}
\author[3]{Nysret Musliu}
\address[3]{TU Wien, Vienna, Austria}

\maketitle
\begin{abstract}
Large language models (LLMs) achieve astonishing results on a wide range of tasks.
However, their formal reasoning ability still lags behind.
A promising approach is Neurosymbolic LLM reasoning.
It works by using LLMs as translators from natural to formal languages
and symbolic solvers for deriving correct results.
Still, the contributing factors to the success of Neurosymbolic LLM reasoning remain unclear.
%
This paper demonstrates that one previously overlooked factor is the choice of the formal language.
We introduce the intermediate language challenge: selecting a suitable formal language for neurosymbolic reasoning.
By comparing four formal languages across three datasets and seven LLMs, we show that the choice of formal language affects both syntactic and semantic reasoning capabilities.
    We also discuss the varying effects across different LLMs.

\end{abstract}


\vspace{-1cm}
\section{Introduction}
\label{sec:introduction}
\vspace{-0.2cm}
Logical reasoning tasks pose a challenge to Large Language Models (LLMs), 
as they struggle to reason abstractly and correctly~\cite{saparov_language_2023,lampinen_language_2024,panas_can_2024}.
This leads to their sometimes spectacular failures, like deriving that birds have four legs~\cite{lin_birds_2020}.
%
%
%
%
One attempt to improve the abstract reasoning capability is Chain of Thought (CoT)~\citep{wei_chain--thought_2022} prompting.
With CoT, LLMs are nudged to reason step-by-step.
However, LLMs' step-by-step reasoning is generally \emph{non-faithful} - even when all individual reasoning steps are correct on their own,
the final conclusion can be false~\citep{lyu_faithful_2023}.
%

\emph{Neurosymbolic LLM reasoning} enables faithful reasoning chains.
It works in two steps:
the first step translates a natural language-posed logical reasoning problem 
into a \emph{formal intermediate language}.
The translation uses the \emph{in-context-learning} (ICL) capability of LLMs.
The second step is to solve the translated problem by a symbolic reasoner.
%
Novel neurosymbolic approaches,
such as Logic-LM~\cite{pan_logic-lm_2023} and LINC~\cite{olausson_linc_2023},
report substantial improvements over pure LLM prompting.

However, it remains unclear what the reasons for their reported success are.
This comes, as there are a plethora of possible contributing factors,
ranging from the LLM training data,
over auxiliary systems (such as re-prompting on errors),
to the choice of formal language.
We investigate the choice of formal language, as it is rarely justified, let alone supported by empirical evidence, leaving its impact on neurosymbolic LLM reasoning largely uncharted.


\noindent
\textbf{Contributions}.
By measuring the impact of different formal languages,
we take a first step toward better understanding why neurosymbolic systems obtain state-of-the-art results
and how the choice of formal language affects reasoning.
The main contributions of this work are:
\vspace{-0.3em}
\begin{itemize}[leftmargin=*]
    \item We introduce the \textit{intermediate language challenge}:
        the choice of formal language for neurosymbolic LLM reasoning affects the reasoning performance.
        \vspace{-.6em}
\item We conduct an extensive empirical study of four formal languages across three logical reasoning datasets (ProntoQA, ProofWriter, FOLIO) and seven LLMs (8B--671B).
\vspace{-0.6em}
%
\vspace{-0.3em}
\end{itemize}
Our experiments show that the choice of formal language matters:
first-order logic outperforms logic programming languages.
This paper is the short version of
\emph{Intermediate Languages Matter: Formal Choice Drives Neurosymbolic LLM Reasoning}~\cite{beiser2025intermediate}.
Here, we focus on the main results of the comparison of different formal languages,
and (previously not shown) comparison of LLMs.

\noindent \textbf{Structure}.
After this introduction we will present the necessary preliminaries and discuss related work (Section~\ref{sec:background}).
We continue to introduce our main hypothesis - the intermediate language problem - that the choice of formal language affects reasoning performance (Section~\ref{sec:the-intermediate-language-problem}),
which we follow with our experimental setup (Section~\ref{section:experimental-scenarios}), and
our experimental results (Section~\ref{section:experiments:experiments}).
We close our paper with our conclusions in Section~\ref{sec:discussion}.

\vspace{-0.4cm}
\section{Preliminaries}
\label{sec:background}
\vspace{-0.2cm}
We briefly present the necessary background material and definitions for understanding the paper.
Recall that the main objective of this study is to compare the reasoning performance of different formal languages on modern LLMs.
By taking the perspective of end-users,
we treat LLMs as immutable black-box next-token predictor machines.
Therefore, we are primarily interested in what effects different prompting strategies have on the reasoning performance
and consider the effects of other techniques, such as fine-tuning, as out of scope.
Throughout this paper, the terms \emph{syntax} and \emph{semantics} are used in their formal language sense.
%
%
%
%
%
%
\subsection{Chain-of-Thought (CoT) prompting}

Chain-of-Thought (CoT) prompting is an \textit{in-context-learning} (ICL)
technique which improves the reasoning capabilities of LLMs by adding additional information to a prompt~\citep{wei_chain--thought_2022}.
CoT nudges the LLM to mimic a reasoning chain,
where we show an example in the next listing.
%
%
\begin{lstlisting}
The following example showcases the line of reasoning you have to follow:
---- Question ----
Each cat is a carnivore. Fae is a cat.
True or false: Fae is a carnivore
---- Reasoning ----
Fae is a cat. Each cat is a carnivore. So Fae is a carnivore.
\end{lstlisting}\vspace{-0.2cm}
%
%

\begin{figure}[t]
    \includegraphics[width=14.5cm]{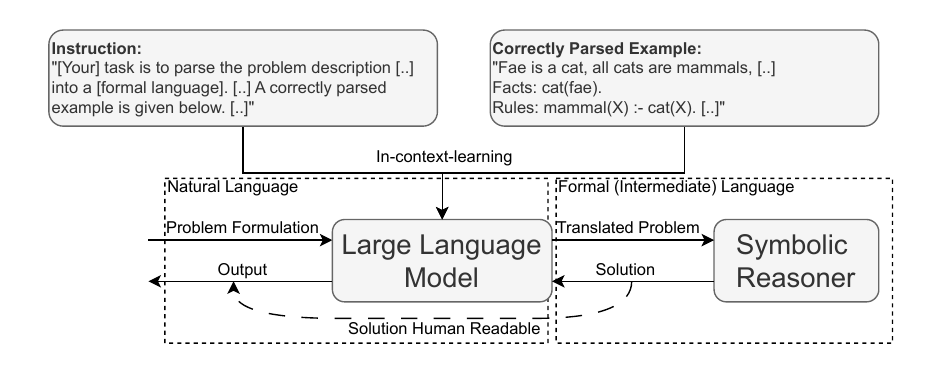}
    \vspace{-1.4em}
    \caption{
        Neurosymbolic LLM reasoning: 
        A problem formulated in natural language is translated by using in-context-learning
        into a formal language.
        Subsequently, a symbolic reasoner subsequently computes a solution to the problem, 
        which is followed by the re-translation of the solution.
    }
    \label{sec:logic_lm:neurosymbolic_schematics}
    \vspace{-1em}
\end{figure}

\vspace{-0.2cm}
\subsection{Neurosymbolic LLM Reasoning}
\label{sec:logic-lm}
\vspace{-0.2cm}
%
Figure~\ref{sec:logic_lm:neurosymbolic_schematics} depicts the high-level schematics of neurosymbolic LLM reasoning.
%
%
%
A natural language-posed problem is translated into its \textit{formal language} representation by using ICL.
ICL comprises of an \emph{ICL-instruction} and an \emph{ICL-example}.
The instruction describes the general task,
while the example showcases how to translate the natural language-posed problem into a formal language.
We refer to the formal language of the ICL-example,
as the \textit{chosen} formal language.

In a second step, the symbolic reasoner solves the problem by obtaining a solution from the formal representation,
which can be either re-translated into natural language or directly used as output.
We do not employ backup strategies and use a as close as possible deterministic prompting (temperature $0$),
as we are interested in the unfiltered affect of the formal language on reasoning performance.
We thereby differ from other related approaches like Logic-LM~\citep{pan_logic-lm_2023}, Logic-LM++~\citep{kirtania_logic-lm_2024}, and LINC~\cite{olausson_linc_2023}.

\vspace{-0.2cm}
\subsection{Related Work}
\label{sec:related-work}
\vspace{-0.2cm}

Improving LLM's reasoning capability was approached by different angles.
Model improvements include fine-tuning or pre-training to improve numerical capabilities~\citep{geva_injecting_2020} or syntax recognition of ASP with LLASP~\citep{coppolillo_llasp_2024}.
Prompting techniques are widely used, such as CoT prompting,
part of the emergent ICL or \textit{few-shot-learning} capability~\citep{shanahan_talking_2024}.
CoT improves LLMs' performance on reasoning tasks~\citep{wei_chain--thought_2022}.
Recent reasoning-focused LLMs, like DeepSeek-R1 utilize internal CoT~\cite{algorithms_of_thought}.
In contrast to these approaches we utilize CoT prompting and neurosymbolic AI.
%
    %
Neurosymbolic AI~\citep{garcez_neurosymbolic_2023} is a broad field which ranges from
differentiable logic~\citep{badreddine_logic_2022}
over visual question answering~\citep{eiter_logic-based_2023},
to LLMs~\citep{pan_logic-lm_2023, olausson_linc_2023}.
For logical reasoning tasks in particular, 
\textit{Logic-LM}~\citep{pan_logic-lm_2023} and \textit{LINC}~\citep{olausson_linc_2023} are two proposed
neurosymbolic approaches that combine LLMs with symbolic solvers.
They translate a natural languages into a formal language - called autoformalization~\citep{wu_autoformalization_2022,liu_how_2024}.
Logic-of-Thought, which tackles logic-puzzles with a neurosymbolic approach~\citep{li2025logicofthoughtempoweringlargelanguage}, is also related.
Although prior work employs an intermediate language, it seldom justifies the choice.
We show — empirically — that the selected language decisively shapes reasoning performance.

\section{The Intermediate Language Challenge}
\label{sec:the-intermediate-language-problem}
\vspace{-0.3cm}

We proceed to define our intermediate language challenge for neurosymbolic LLM reasoning.
We assume to have given a natural language-posed reasoning problem $\mathcal{P}$ and
a set of possible formal languages $\mathcal{L}$.
\vspace{-.3cm}
\begin{definition}
    The intermediate language challenge is the task of
    choosing a formal language $l \in \mathcal{L}$ for solving $\mathcal{P}$ with a high 
    reasoning accuracy.
\end{definition}
\vspace{-.3cm}
Inherent to the intermediate language challenge is autoformalization~\citep{wu_autoformalization_2022}.
\vspace{-.3cm}
\begin{definition}
Let $l \in \mathcal{L}$ be a fixed formal language.
Then, autoformalization aims for automatic and correct translation of $\mathcal{P}$ into $l$.
\end{definition}
\vspace{-.3cm}
While autoformalization is concerned with the correct translation from natural language into a fixed formal language $l$,
the intermediate language challenge is about choosing a suitable formal language $l' \in \mathcal{L}$ s.t. autoformalization can be done effectively.
%
%
%
We identify two root causes of the intermediate language problem:
%
%
(i) Syntax affects LLMs' reasoning performance, and (ii) one logical problem can be translated into multiple formal languages.

\noindent
\textbf{Syntax affects LLMs' reasoning performance}.
Consider the following two logical reasoning problems:
(1) ``Tommi is a tumpus. Each tumpus is a wumpus. Is Tommi a wumpus?''
(2) ``Tommi is a cat. Each cat is an animal. Is Tommi an animal?''
Recent work suggests that, on average, LLMs perform better for scenarios of type (2) than type (1)~\citep{saparov_language_2023,lampinen_language_2024}.
From a semantic perspective, both scenarios require the application of modus ponens.
Thus, as the \emph{only} difference lies in the \emph{syntax}, we can conclude that the syntax affects LLMs' reasoning capabilities.
Going back to formal languages,
observe that the syntax of formal languages differs.
Therefore, we conclude that the choice of formal language affects LLMs' reasoning capabilities.

%
%

\noindent
\textbf{Logical problems can be encoded in different formal languages}.
Take the logical reasoning problem (2) from the paragraph above. 
This problem can be encoded in different formal languages, such as logic programming or first-order logic (FOL),
while maintaining semantic correctness.
%
%
%

\vspace{-0.4cm}
\section{Experiment Setup}
\vspace{-0.2cm}
\label{section:experimental-scenarios}
To show the impact of the intermediate language challenge, we investigate a set of formal languages $\mathcal{L} = \{\text{Pyke},\text{ASP},\text{NLTK},\text{FOL}\}$.
We conduct experiments on three different datasets, ProntoQA~\cite{saparov_language_2023}, ProofWriter~\cite{tafjord_proofwriter_2021}, and FOLIO~\cite{han_folio_2024}.
Let $\mathcal{D}$ be a given dataset, then each data instance $\mathcal{P} \in \mathcal{D}$ can be considered a reasoning problem.
Each $\mathcal{P}$ is translated into a formal language by the LLM according to a specification (prompting style).
We prompt the LLM with a prompting style that adheres to Figure~\ref{sec:logic_lm:neurosymbolic_schematics}
- i.e., we provide an ICL instruction and an ICL example.
We use a set of prompting styles, where they differ in the syntax of the ICL example, such as wrapping the example in markdown syntax.
Importantly, we enable comparability between formal languages by using the same set of prompting styles for each formal language.

\vspace{-0.3cm}
\subsection{Formal Languages}
\vspace{-0.2cm}

We will provide a brief overview of the formal languages $\mathcal{L}$ used for our experiments.

\noindent \textbf{Pyke}:
The logic programming derivative Pyke~\citep{frederiksen_applying_2008}
expresses rules similar to \textit{if}-\textit{then} statements.
Pyke derives 
conclusions by forward, or backward chaining algorithms.

\noindent \textbf{ASP}:
In the non-monotonic logic programming paradigm Answer Set Programming (ASP)~\citep{gelfond_logic_2002,schaub_special_2018} a program is written as a set of rules, which is first grounded~\cite{kaminski_foundations_2023} and then solved~\cite{gebser_theory_2016}.
%

\noindent \textbf{NLTK}:
The natural language toolkit~\citep{bird_natural_2009} is a Python library that enables an integration
of FOL with Prover9~\citep{mccune_prover9_2010}.
We assume familiarity with the semantics of FOL.
%

\noindent \textbf{FOL}:
We assume familiarity with the syntax and semantics of FOL.
For our experiments, we implemented a \textit{parser} that translates FOL to NLTK formulas, which are then solved by Prover9.

\subsection{Datasets}
\vspace{-0.2cm}
We perform experiments on three datasets.
We used one partly hand-crafted ICL-example (training data) per dataset/formal language,
which is not part of the test set.
Each test set configuration resembles the configuration of Logic-LM.
%

\noindent \textbf{ProntoQA}~\citep{saparov_language_2023}.
The ProntoQA dataset is a generated dataset.
We use the fictional character version
with a reasoning depth of 5. 
A random answer has a probability of $50\%$ for getting a correct answer (closed-world assumption - CWA), and a test set with 500 samples is used.

\noindent \textbf{ProofWriter}~\citep{tafjord_proofwriter_2021}.
ProofWriter is a generated dataset.
We chose a reasoning depth of 5.
A random answer has a probability of about $33\%$ to get a correct answer (open-world assumption - OWA).
The test set has 600 samples.

\noindent \textbf{FOLIO}~\citep{han_folio_2024}.
FOLIO is a (partly) expert-written dataset.
A random answer is correct with about $33\%$ (OWA).
The FOLIO test set has 204 samples.
We do not use ASP and Pyke on FOLIO, as FOLIO instances require classical logic concepts which are effectively impossible to encode in standard logic programming.
%

\vspace{-0.3cm}
\subsection{Large Language Models}
\vspace{-0.2cm}

We compare the formal languages on seven LLMs, ranging from $8$B to $671$B parameters.
For all experiments we set the temperature to 0, to obtain a near-deterministic behavior.
We restricted the maximum number of new tokens to be 2048
and did not perform any additional modifications to the LLMs.
%
%
We are primarily interested in how the intermediate language affects small language models (SLMs) with $\approx$ $8$B parameters,
due to their lower resource consumption.
Further, we focus on chat models, as reasoning models build upon them.
%
%
We used the following LLMs of approximately 8B parameters:
\textit{GPT-4o-mini}\footnote{\url{https://platform.openai.com/docs/models/gpt-4o-mini}},
\textit{Ministral-8B}\footnote{\url{https://mistral.ai/news/ministraux}},
\textit{Llama-8B}\footnote{\url{https://openrouter.ai/meta-llama/llama-3.1-8b-instruct}}.
and \textit{DeepSeek-8B}\footnote{\url{https://openrouter.ai/deepseek/deepseek-r1-distill-llama-8b}}.
To study the effects when using bigger models,
we additionally perform experiments on 
\textit{DeepSeek-32B}\footnote{\url{https://openrouter.ai/deepseek/deepseek-r1-distill-qwen-32b}} ($\approx$ $32$B parameters)
and \textit{DeepSeek-V3} ($\approx$ $671$B parameters) models.
To verify the results on state-of-the-art reasoning models
we performed benchmarks on \textit{DeepSeek-R1}\footnote{\url{https://api-docs.deepseek.com/news/news1226}} as well.
We prompted \textit{DeepSeek-R1} with both 2048 and 20480 max-output-tokens, due to increased output token generation of the reasoning model.

\vspace{-0.3cm}
\subsection{Baselines}
\vspace{-0.2cm}
The \textit{chance} is the probability of getting a correct answer by a random draw.
Chance is $50\%$ for ProntoQA, as it has a CWA,
and $33\%$ for ProofWriter and FOLIO, as they have an OWA.
Additionally, we use the following baselines\footnote{We additionally show the results of Neurosymbolic baselines in the main paper.}.

\noindent \textbf{Std.} - 
refers to standard prompting.
The LLM is given a short instruction on the task,
the natural language-posed problem,
and a short example of how the LLM shall answer the question.

\noindent \textbf{CoT} - 
refers to CoT prompting.
It extends standard prompting by nudging the LLM to reason step-by-step by employing CoT.


\vspace{-0.2cm}
\subsection{Experimental Evaluation}
\vspace{-0.2cm}

\begin{figure}
    \begin{subfigure}{0.5\textwidth}
        \includegraphics[width=7.3cm]{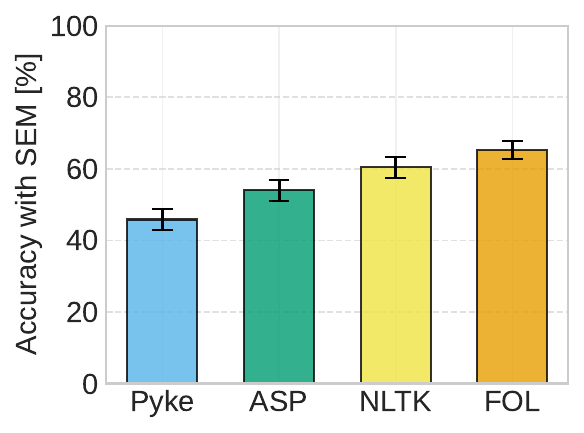}

    \end{subfigure}
    \begin{subfigure}{0.49\textwidth}
        \includegraphics[width=7.3cm]{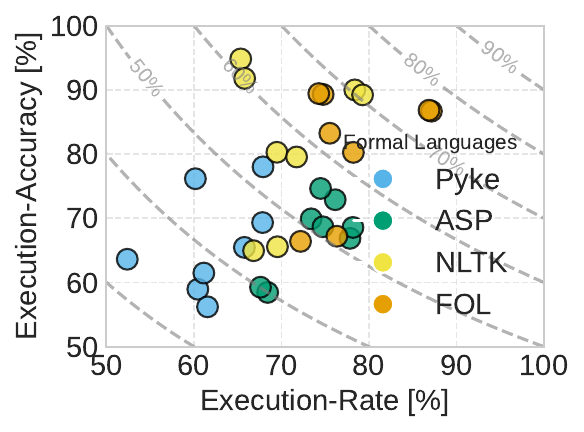}
    \end{subfigure}
    \caption{
        Left: We show the effects of the formal languages, averaged across all prompting styles, LLMs, and the ProntoQA and ProofWriter datasets.
    Error bars show the SEM $n=112$.
    Right: Scatter plots comparing execution-rate to execution-accuracy for the formal languages.
    A single dot shows an average across prompting styles averaged over ProntoQA and ProofWriter datasets and all LLMs ($n=14$).
    Contour lines show overall-accuracy in steps of $10\%$.
    }
    \label{fig:max-results-folio-ablation}
    \vspace{-0.5cm}
\end{figure}

%
We conduct our experiments on an adapted Logic-LM implementation.
Our adaptation includes an ASP symbolic solver based on Clingo~\citep{gebser_theory_2016},
a new Pyke implementation, and an adapted NLTK/FOL solver implementation.
%
We conduct experiments for $4$ formal languages and $8$ prompting styles, leading to $32$ total experiments for ProntoQA and ProofWriter.
Including the $4$ baseline experiments, we report $36$ experiments, respectively.
For FOLIO, we conduct $20$ experiments in total (Pyke and ASP cannot be measured).
This leads to a total of $92$ experiments per LLM and
$644$
experiments in total.
The overall number of queries is $43680$ per LLM and overall $305760$.
Let $\#D$ be the dataset size,
$\#\text{EXEC}$ the number of correctly parsed instances, and
$\#\text{TRUE}$ the number of correctly solved instances.
\emph{Syntactically correct} refers to a translation that adheres to the defined formal language,
whereas \emph{correctly solved} refers to a correct syntactical translation and the correct output of the solver.
%
%
The \textit{execution-rate} is the fraction of correct syntactical outputs (Exec-Rate, $\frac{\#\text{EXEC}}{\#D}$),
\textit{execution-accuracy}, is the fraction of correctly solved instances of all syntactically correct ones (Exec-Acc, $\frac{\#\text{TRUE}}{\#\text{EXEC}}$),
and \textit{overall-accuracy} 
is the fraction of correctly solved instances over the entire dataset (Overall-Acc, $\frac{\#\text{TRUE}}{\#\text{\#D}}$).
Observe: $\textit{Overall-Acc} = \textit{Exec-Acc} \cdot \textit{Exec-Rate}$.
%
Baselines which do not use neurosymbolic reasoning are considered to have an execution-rate of $100\%$,
while their execution-accuracy resembles their overall-accuracy,
as they are not required to adhere to a formal language.

\vspace{-0.2cm}
\section{Results}
\label{section:experiments:experiments}
\vspace{-0.2cm}

\begin{table}
    \centering
{\scriptsize
    \begin{tabular}{c||c|c|c||c|c|c||c|c|c}
        Lang. & \multicolumn{3}{c}{Overall Results} & \multicolumn{3}{c}{GPT-4o-mini} & \multicolumn{3}{c}{Ministral-8B}  \\ \toprule
         & Avg. & \multicolumn{2}{l}{SEM} & Avg. & SEM & Max & Avg. & SEM & Max \\ \midrule
        Std. & / & \multicolumn{2}{l}{/} & / & / & 70.20 & / & / & 48.80  \\\midrule
        CoT     & / & \multicolumn{2}{l}{/} & / & / & 84.00 & / & / & 86.60  \\\midrule
        Pyke & 45.83 & \multicolumn{2}{l}{2.82} & 59.76 & 4.56 & 93.80 & 39.93 & 5.15 & 69.60  \\\midrule
        ASP & 53.94 & \multicolumn{2}{l}{2.94} & 61.00 & 4.17 & 97.20 & 34.98 & 3.42 & 61.40  \\ \midrule
        NLTK & 60.36 & \multicolumn{2}{l}{2.92} & 72.74 & 5.42 & 99.80 & 75.20 & 6.40 & 99.60 \\\midrule
        FOL & \textbf{65.29} & \multicolumn{2}{l}{2.52} & \textbf{72.85} & 4.75 & \textbf{100.00} & \textbf{75.94} & 6.40 & \textbf{100.00} \\ \midrule
        Lang. & \multicolumn{3}{c}{Llama-8B} & \multicolumn{3}{c}{DeepSeek-8B} & \multicolumn{3}{c}{DeepSeek-32B}  \\ \toprule
         & Avg. & SEM & Max & Avg. & SEM & Max & Avg. & SEM & Max  \\ \midrule
        Std. & / & / & 52.00 & / & / & \textbf{87.40} & / & / & \textbf{99.20}  \\\midrule
        CoT     &/ & / & 68.80 & / & / & \textbf{87.40} & / & / & 98.80  \\\midrule
        Pyke & 20.76 & 5.43 & 61.00 & 1.43 & 0.48 & 5.80 & 53.44 & 3.48 & 79.40  \\\midrule
        ASP & 6.78 & 1.15 & 19.33 & 23.98 & 4.14 & 46.40 & \textbf{63.74} & 1.94 & 77.60 \\ \midrule
        NLTK & \textbf{54.94} & 5.76 & \textbf{93.20} & 15.77 & 2.13 & 29.00 & 59.44 & 5.58 & 96.00  \\\midrule
        FOL & 51.97 & 4.74 & 77.00 & \textbf{33.15} & 4.58 & 69.80 & 61.11 & 5.67 & 87.00 \\ \midrule
        Lang. & \multicolumn{3}{c}{DeepSeek-V3} & \multicolumn{3}{c}{DeepSeek-R1} & \multicolumn{3}{c}{DeepSeek-R1 (20480)}  \\ \toprule
         & Avg. & SEM & Max & Avg. & SEM & Max & Avg. & SEM & Max  \\ \midrule
        Std. & / & / & 98.00 & / & / & 57.60 & / & / & 97.40  \\\midrule
        CoT     &/ & / & 99.80 & / & / & \textbf{81.80} & / & / & 99.00  \\\midrule
        Pyke & 68.20 & 2.94 & 82.40 & 12.88 & 4.93 & 73.17 & 77.28 & 4.76 & 98.00  \\\midrule
        ASP & 78.32 & 5.80 & 99.00 & \textbf{48.17} & 3.09 & {74.00} & \textbf{88.82} & 1.81 & 98.40 \\ \midrule
        NLTK & \textbf{84.07} & 5.40 & \textbf{100.00} & 20.89 & 3.89 & 46.67 & 80.33 & 5.52 & 98.40  \\\midrule
        FOL & 76.54 & 6.57 & \textbf{100.00} & 21.09 & 4.11 & 43.17 & 85.45 & 4.12 & \textbf{99.80} \\ \midrule
    \end{tabular}
    }
    \vspace{-0.5cm}
    \caption{Overall and per LLM results.
    LLMs prompted with temperature 0 and max-output-tokens $2048$, except for DeepSeek-R1 (20480).
    All values in $[\%]$.
    For overall results Avg. and SEM, $n=112$. 
    For per LLM result: $n=16$ for the neurosymbolic approaches and $n=2$ for the baselines.
    }
    \label{tab:ablations-study-scenarios-per-formal-language}
\vspace{-0.5cm}
\end{table}

We show the experimental results in Figure~\ref{fig:max-results-folio-ablation} and Table~\ref{tab:ablations-study-scenarios-per-formal-language}.
In this paper we focus on the main results of the formal languages and provide further information on reasoning model performance and averaged LLM performance.
Further results are shown in the main paper, such as an ablation study on eight different prompting styles and how well formal languages work on each dataset.
Overall we report that FOL achieves the best results,
followed by NLTK, ASP, and lastly Pyke.
We report these findings in Figure~\ref{fig:max-results-folio-ablation} (left) and Table~\ref{tab:ablations-study-scenarios-per-formal-language} (left top), where we show averaged results with the standard error of the mean (SEM).
For averaging the formal languages, we compute the average across all LLMs, prompting styles,
and the datasets ProntoQA and ProofWriter, leading to $n=112$.
To account for a fair comparison and incorporation of the results of the reasoning model,
we only used the 20480 token results for DeepSeek-R1 for these averages.
For the problems in the datasets, we do not encounter difficulties when solving in terms of \emph{intractability} - a combinatorial explosion in the solver - we never exceed 60s computation time.
Therefore, we are not required to use special strategies for tackling intractability,
such as symmetry breaking~\cite{fahle_symmetry_2001} or tackling the ASP bottleneck~\cite{beiser_bypassing_2024}.
In Figure~\ref{fig:max-results-folio-ablation} (right), we show averaged scatter plots of the execution-rate (x-axis) vs. execution-accuracy (y-axis).
Each dot represents a formal language with a specific prompting style,
averaged across all LLMs and ProntoQA, and ProofWriter datasets ($n=14$).
The overall-accuracy is obtained by multiplying a point's x-position with its respective y-position.
We report that Pyke performs approximately equally well on execution-rate and execution-accuracy,
while ASP's execution-rate tends to stay relatively high.
Further, NLTK's execution-accuracy is relatively high,
as it stays above $60\%$. 
FOL's behavior is similar to NLTK's.
Still FOL has a higher execution-rate, resulting in a higher overall-accuracy.

In Table~\ref{tab:ablations-study-scenarios-per-formal-language} (all except left top) we show the individual
results of the formal languages on the LLMs.
We report that the results differ widely between LLMs.
Considering the average case, FOL achieves the highest results on three (GPT-4o-mini, Ministral-8B, and DeepSeek-8B), NLTK on two (Llama-8B and DeepSeek-V3) and ASP on three (DeepSeek-32B, DeepSeek-R1, and DeepSeek-R1 (20480)) LLMs.
However, these values are often in the range of the SEM, therefore, inconclusive.
For example, on GPT-4o-mini FOL has $72.85\%$ and an SEM of $4.75\%$,
whereas NLTK has $72.74\%$ and an SEM of $5.42\%$.
Regarding the best results, FOL achieves on four (GPT-4o-mini, Ministral-8B, DeepSeek-V3, and DeepSeek-R1 (20480)),
NLTK on two (Llama-8B and DeepSeek-V3), and the baselines on three (DeepSeek-8B, DeepSeeek-32B, and DeepSeek-R1) LLMs.
We report the largest performance improvements on SLMs,
such as GPT-4o-mini, Ministral-8B, and Llama-8B.
However, on SLMs also the performance fluctuates the most.
For example, ASP does not perform well on Llama-8B, whereas Pyke does not perform well on DeepSeek-8B.
The reasoning model DeepSeek-R1 needs more output tokens to generate suitable responses,
when compared to the chat models.
Although not a surprise, this has consequences.
While for all chat LLMs a max-output-token size of 2048 is sufficient, 
for DeepSeek-R1 only a max-output-token size of 20480 yields reliably non-truncated results.

\vspace{-0.2cm}
\section{Conclusion and Discussion}
\label{sec:discussion}
\vspace{-0.2cm}
Logical reasoning tasks pose a problem to LLMs,
as they remain limited in their ability to perform probabilistic retrieval~\cite{panas_can_2024}.
Neurosymbolic approaches help,
by constraining the probabilistic nature to the translation step of a 
natural language-posed problem into a formal language~\citep{pan_logic-lm_2023,olausson_linc_2023}.
Therefore, the reasoning step itself is not affected by the probabilistic nature of LLMs.

In this paper, we discuss the effect of the chosen formal language on a model's reasoning performance.
We introduce the \textit{intermediate language challenge},
which refers to the problem of choosing a suitable formal language for neurosymbolic reasoning.
In our experiments, we compare Pyke, ASP, NLTK, and FOL as formal languages.
The results show that FOL performs best, followed by NLTK and ASP,
with Pyke coming last.
When analyzing the behavior of different formal languages we observed translation errors which were unique to the formal language, and errors common across different formal languages.
For \emph{Pyke} in particular, we notice that LLMs format the output incorrectly, by missing line breaks.
When translating to \emph{ASP}, LLMs struggle to distinguish the two notions of negation:
\emph{strong}, written as $-$, and \emph{default}, written as $not$.
This results in program statements such as \textit{-not p1(wren)}, which are syntactically incorrect.
The syntactic errors between \emph{NLTK} and \emph{FOL} are similar.
Examples include incorrectly setting parentheses or using a predicate with multiple arities - e.g., $p14(X)$ and $p14(X,Y)$. 

Our results between different LLMs vary widely and suggest that for neurosymbolic LLM reasoning the usage of huge LLMs is not always justified,
as for our tasks we already achieved $100\%$ max-accuracy on GPT-4o-mini and Ministral-8B (8B models).
Further, while for all chat models max-output-tokens of $2048$ was sufficient,
this was not the case for DeepSeek-R1, where we needed more (we increased the parameter to $20480$).
Interestingly, our results indicate that ASP uses less output tokens,
as it achieved decent results on DeepSeek-R1 (2048).
Further, we observed the largest improvements w.r.t. to the baselines and the largest variations in performance on small language models (SLMs).
We hypothesize that the performance differences can be explained with a lack or abundance of the formal languages in the training data.
This opens the door for crafting custom intermediate languages and fine-tuning the SLMs on these custom languages, which we plan to explore in a future study.

%
%



\clearpage

\section*{Declaration on Generative AI}
\vspace{-0.2cm}

During the preparation of this work, the authors used OpenAI O3 and Grammarly in order to: Grammar and spelling check. 
The seven LLMs
(\textit{GPT-4o-mini}, \textit{Ministral-8B}, \textit{Llama-8B},
\textit{DeepSeek-8B}, \textit{DeepSeek-32B}, \textit{DeepSeek-V3} and \textit{DeepSeek-R1})
were used, as discussed in Section~\ref{section:experimental-scenarios}.
\vspace{-0.5cm}

\section*{Acknowledgements}
\vspace{-0.2cm}

This research was supported by Frequentis,
FWF grant 10.55776/COE12,
and the Dieberger-Peter Skalicky Stipend.
\vspace{-0.5cm}

%
%
\bibliography{sample-ceur}


\end{document}